\documentclass[runningheads]{llncs}

\usepackage{eccv}

\usepackage{eccvabbrv}
\usepackage{enumitem}
\usepackage{graphicx}
\usepackage{booktabs}
\usepackage[accsupp]{axessibility}  

\usepackage{hyperref}

\usepackage{orcidlink}
\usepackage[textsize=tiny]{todonotes}
\usepackage{bibentry}
\usepackage{amssymb} 
\usepackage{amsmath}
\usepackage{booktabs} 
\usepackage{multirow}
\usepackage[table]{xcolor}

\begin{document}

\title{Circuit-MLLM: Topological Logic-Guided Latent-Space Visual Reasoning for Circuit Schematic Understanding} 

\titlerunning{Circuit-MLLM}

\author{Jinyuan Deng \and
Yuqi Jiang \and
Wenjing Huang \and
Xin Li \and
Qi Sun\thanks{Corresponding author.} \and
Cheng Zhuo} 

\authorrunning{J.~Deng et al.}

\institute{Zhejiang University, Hangzhou, China\\
\email{qisunchn@zju.edu.cn}}

\maketitle

\begin{abstract}
Through pre-training on extensive text and image datasets, current multi-modal large language models (MLLMs) achieve strong performance on general tasks. 
However, circuit schematics present a unique challenge for MLLMs due to their dense component layouts and distinct topological logic, demanding fine-grained structural parsing to extract the electrical semantics.
To address this, we propose Circuit-MLLM, a multimodal reasoning framework that reformulates circuit topology analysis as a process of device localization, path tracing, and sequential reasoning within the latent space.
We introduce a circuit knowledge mining mechanism that deeply aligns the model's latent representations with structurally rich features derived from multi-granularity circuit vision experts, enabling the model to effectively internalize topological semantics.
Building upon these internalized semantics, we devise a topology-guided sequencing strategy that decouples reasoning from the rigid raster-scan order, enforcing stepwise inference along the circuit's topological logic in latent space.
Across diverse circuit analysis tasks, Circuit-MLLM consistently outperforms strong baselines, notably achieving a 25\% higher average score than GPT-5.1, which demonstrates the effectiveness of our framework in circuit schematic topology analysis.
Code is publicly available at https://github.com/IC-Yuan/Circuit-MLLM.
  \keywords{Multimodal Reasoning \and Latent-space \and Circuit Schematics}
\end{abstract}
\section{Introduction}
Recent advances in Multimodal Large Language Models (MLLMs) have shown strong performance on general vision–language understanding and reasoning \cite{li2022blip,liu2023visual}. 
However, applying MLLMs to circuit schematics remains challenging due to their dense component layouts and intricate topological dependencies \cite{yue2024mmmu}. 
This limitation impedes the deployment of MLLMs in Electronic Design Automation (EDA). 
Consequently, empowering MLLMs with schematic parsing capabilities is pivotal for automating circuit analysis and accelerating chip design.

Traditional circuit analysis predominantly depends on rigid computer vision pipelines that employ rule-based algorithms for component and connection recognition \cite{shi2024amsnet}.
Following this paradigm, several MLLM-based approaches for circuit understanding rely on specialized detectors as external front-ends for symbol localization \cite{bhandari2024masala}, adopting a decoupled design that suffers from error propagation and inhibits the MLLM's inherent reasoning capabilities.
In contrast, end-to-end approaches \cite{zhu2025maps} bypass external detectors and typically rely on supervised fine-tuning (SFT) with schematic-to-structured language alignment. 
However, due to the lack of rich instance-level annotations, these models often rely on large-scale synthetic data, which limits their generalization to real-world distributions.

\begin{figure*}[t]
  \centering
  \includegraphics[width=\textwidth]{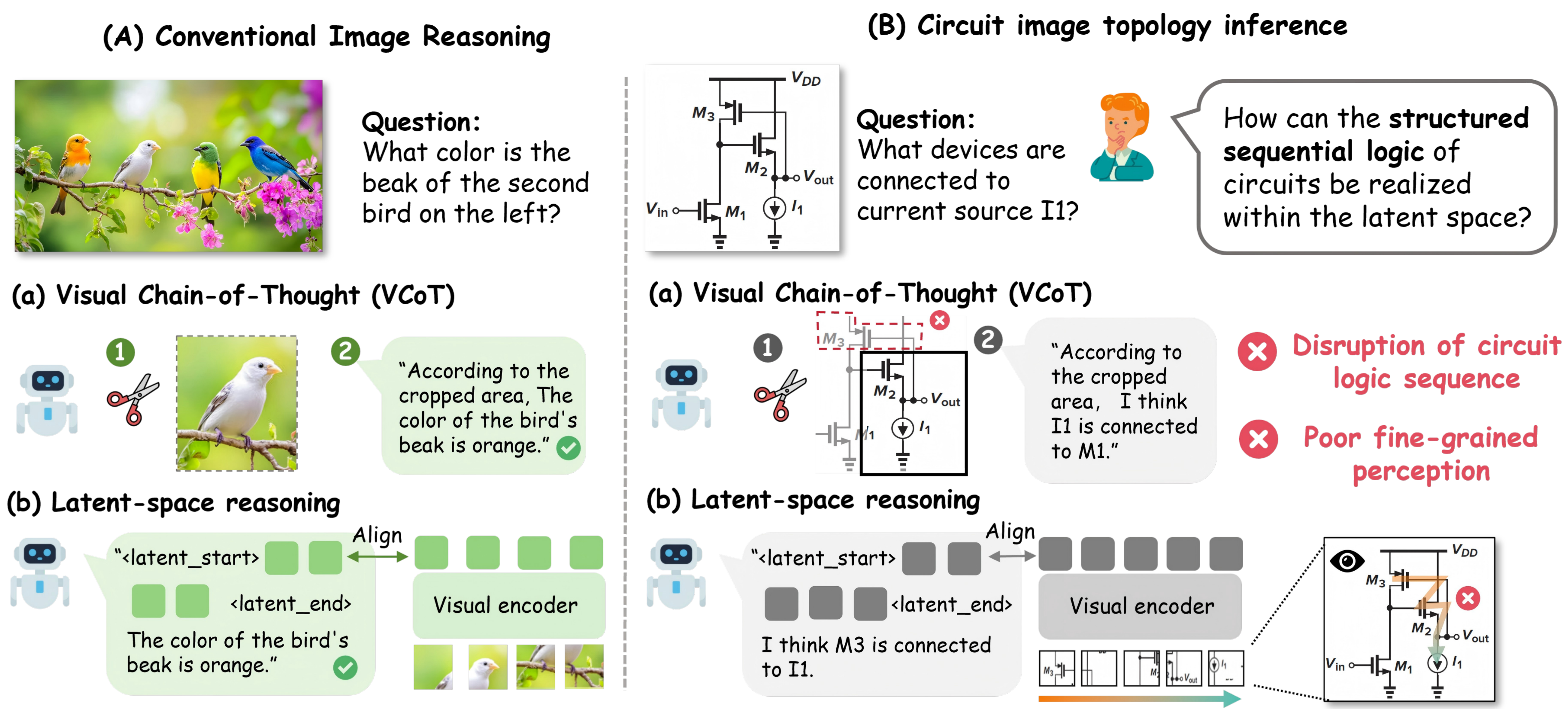} 
  \caption{\textbf{Motivation of Circuit-MLLM.} Existing Visual Chain-of-Thought methods often disrupt the topological integrity of circuits. Meanwhile, latent-space approaches, constrained by raster-order alignment and coarse-grained feature extraction, force the model to deviate from topological paths and result in a lack of fine-grained perception.}
  \label{fig:motivation}
\end{figure*}
Recent advancements in MLLMs often employ Textual Chain-of-Thought to enhance reasoning, attempting to use textual descriptions to guide the model's attention toward specific image regions. For circuit schematics, however, describing topology and routing precisely in natural language is inherently brittle, frequently leading to semantic ambiguity and background distraction.
Visual Chain-of-Thought (VCoT) \cite{zhang2023multimodal} offers a novel perspective by explicitly invoking external tools such as image cropping. Nevertheless, these approaches \cite{shao2024visual,chen2024visual} typically adopt a region-centric “locate-judge” paradigm. This design is fundamentally ill-suited for circuit topology analysis, as discrete bounding boxes often fail to preserve the continuity of thin, long-range wires, resulting in fragmented connections and excessive background clutter.

In contrast, latent-space visual reasoning \cite{li2025latent,yang2025machine,dong2025interleaved} offers an alternative by performing inference within hidden states, avoiding explicit bounding-box and text-based representations. This paradigm provides a distinct advantage for analyzing irregular topologies that are inherently difficult to describe. However, directly applying this latent paradigm to complex circuit topologies faces two key challenges:
\textbf{(1) Scarcity of fine-grained features:} Current latent-space approaches typically capture only coarse, macroscopic topological regions. These latent representations fail to encode the highly fine-grained connection details and dense component layouts essential for circuit topology analysis.
\textbf{(2) Missing structured order:} Existing latent space alignment methods primarily enforce feature-level alignment between latent vectors and auxiliary image patches. Consequently, the sequence of latent representations defaults to a spatial, top-left to bottom-right raster-scan order. However, this scanning order fails to constrain the topology-driven viewpoint progression, resulting in a fundamental misalignment with the topology's intrinsic logic.

To overcome these challenges, we propose Circuit-MLLM, the first topological logic-guided latent-space visual reasoning framework dedicated to complex circuit topology analysis. First, to address the lack of fine-grained circuit feature perception, we introduce a \textbf{circuit knowledge mining mechanism} based on the latent space. By constructing circuit vision experts, we guide the model to deeply internalize and align complex circuit representations directly within the latent space during training. This mechanism prevents the model from merely focusing on superficial, coarse regional information, thereby endowing it with genuine, fine-grained circuit recognition capabilities.
Furthermore, we propose a \textbf{topology-guided sequencing strategy.} This strategy mandates that the generated latent vectors not only accurately encode the topological regions but also strictly follow the intrinsic topological logic in their generation order by specifically progressing from the root component, along the wire topology, to the target component, thereby effectively simulating the topological viewpoint shifts of human engineers entirely within the latent space.

Our contributions are as follows:
\begin{enumerate}
  \item We propose Circuit-MLLM, a novel latent-space visual reasoning paradigm for end-to-end multimodal analysis of circuit topologies. Unlike verbose Chain-of-Thought or external tool-chain methods, our framework directly bridges the semantic gap between spatial visual priors and rigorous circuit logic within the latent space, enabling intrinsic schematic comprehension.
  \item We design a circuit knowledge mining mechanism based on latent space guided by a set of vision experts, which can generate highly discriminative fine-grained circuit latent representations without increasing inference overhead, thereby significantly improving topology recognition accuracy.
  \item We propose a topology-guided sequencing strategy that decouples latent reasoning from rigid spatial raster-scanning. By dynamically aligning the inference trajectory with the inherent circuit structure, it endows the model with native, human-like topological tracing capabilities.
  \item Extensive experiments reveal that Circuit-MLLM achieves a 25\% higher average score than GPT-5.1 across diverse circuit analysis tasks, and outperforms latent-space visual baselines by up to 26\% on topology reasoning tasks. 
\end{enumerate}

\section{Preliminaries}
\subsection{Automated Circuit Schematic Topology Extraction}
\label{sec:Topology Extraction}
\begin{figure*}[t]
  \centering
  \includegraphics[width=\textwidth]{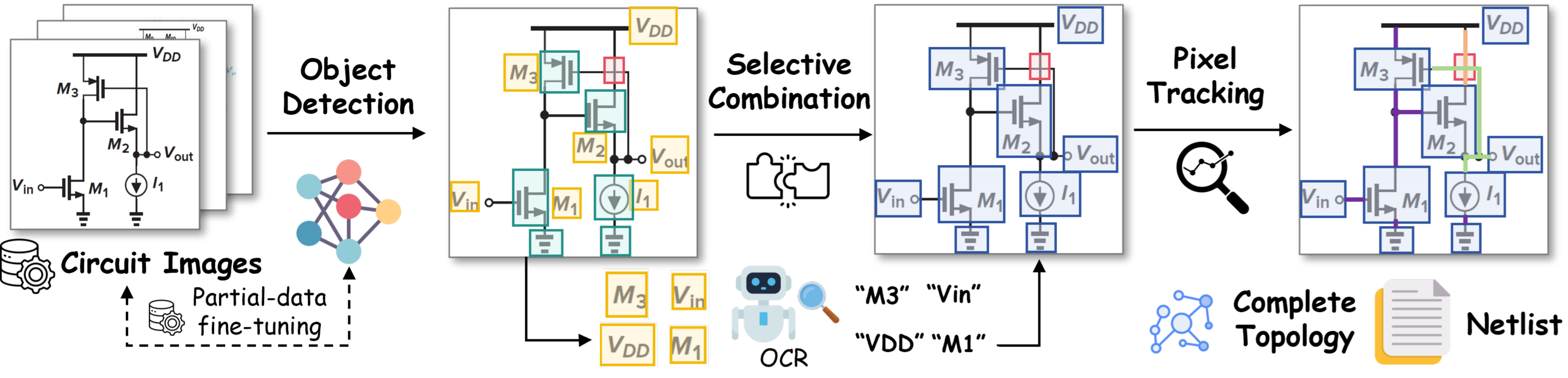} 
  \caption{\textbf{Automated topology annotation workflow for circuit schematics}}
  \label{fig:annotation workflow}
\end{figure*}

We have constructed an automated, fine-grained annotation workflow for circuit schematics as shown in \Cref{fig:annotation workflow}. First, we collected a total of 12,000 circuit schematics from multiple public datasets\cite{gao2025analoggenie,shi2025amsnet,bhandari2024masala,zhu2025maps}, comprehensively covering analog, digital, and mixed-signal systems, making it the largest circuit schematic dataset currently available. Subsequently, we manually annotated a subset of the components, text, and junction nodes. A YOLO11 model \cite{yolo11_ultralytics} was then trained on this subset to enable automated detection and bounding box extraction across the entire dataset. Following this, we introduced an OCR framework \cite{cui2025paddleocr} to perform text recognition and component matching, ensuring the accuracy of attribute assignments. Next, we employed pixel tracking techniques to reconstruct the topological dependencies of the circuits. Given the complexity of circuit topologies and connections, human expert intervention was incorporated to ensure rigorous accuracy. The resulting dataset provides netlist-style descriptions, visual grounding, and pixel-level regional masks. Simultaneously, we constructed latent-space reasoning QA pairs tailored to different question types, and introduced Circuit-MLLM-Bench to evaluate the model's capabilities in circuit topology analysis, as detailed in \Cref{sec:Experiments}.

\subsection{Related Works}
\textbf{Circuit Schematic Comprehension with MLLMs.}
Circuit schematics are a fundamental form of structured visual data in EDA, and their understanding has recently attracted growing attention in the MLLM community \cite{shi2025amsbench,xiang2025seephys}. Prior work mainly follows two paradigms. Tool-based visual reasoning methods (e.g., Masala CHAI\cite{bhandari2024masala}) rely on conventional vision components such as object detectors to produce structured annotations that guide MLLM reasoning. Text-conversion pipelines (e.g., MAPs \cite{zhu2025maps}) first translate circuit schematics into structured netlists using supervised fine-tuned models or off-the-shelf parsers, and subsequently delegate reasoning to text-domain LLMs. However, both paradigms are bottlenecked by intermediate tools and conversions, suffering from error propagation and engineering overhead.

\noindent\textbf{Multimodal Chain-of-Thought.}
Chain of Thought (CoT) prompting unlocked progressive reasoning in large language models via intermediate logical steps \cite{feng2023towards}. Recent studies extended CoT into multimodal settings. For instance, Vision R1 \cite{huang2025vision} deduces complex logic directly from images via multimodal reinforcement learning, while ICoT \cite{zhang2023multimodal} interleaves attention-selected image crops with text tokens to improve visual question answering. Visual CoT \cite{shao2024visual} provides bounding box-grounded reasoning, enhancing spatial localization via explicit visual tokens. Others \cite{hu2024visual} employ external tools for visual evidence, augmenting multimodal CoT. However, these approaches encounter limitations in circuit schematics, where irregular topological regions are ill-suited for standard bounding boxes and challenging to trace via pure natural language.

\noindent\textbf{Latent Visual Reasoning.}
Recent studies have introduced latent space reasoning methods \cite{shen2025codi} into multimodal domains. Specifically, these methods represent intermediate visual steps within the latent space rather than explicitly decoding them into images, thereby preventing the model from being constrained by rigid formats. LVR \cite{li2025latent} and Mirage \cite{yang2025machine} pioneered the Latent Visual Reasoning paradigm, accomplishing latent reasoning by aligning auxiliary image features with vectors in the latent space. Building upon this, ILVR \cite{dong2025interleaved} introduced selective latent vector alignment alongside an interleaved multiple round latent space reasoning paradigm. Despite the unconstrained latent space being highly suitable for complex circuit topologies, current methods often overlook its intrinsic logical chains, causing reasoning misalignments in circuit topology analysis.
\subsection{Visual Attribution of MLLMs in Circuit Analysis}
\label{sec:Visual Attribution}
\begin{figure*}[t]
  \centering
  \includegraphics[width=0.95\textwidth]{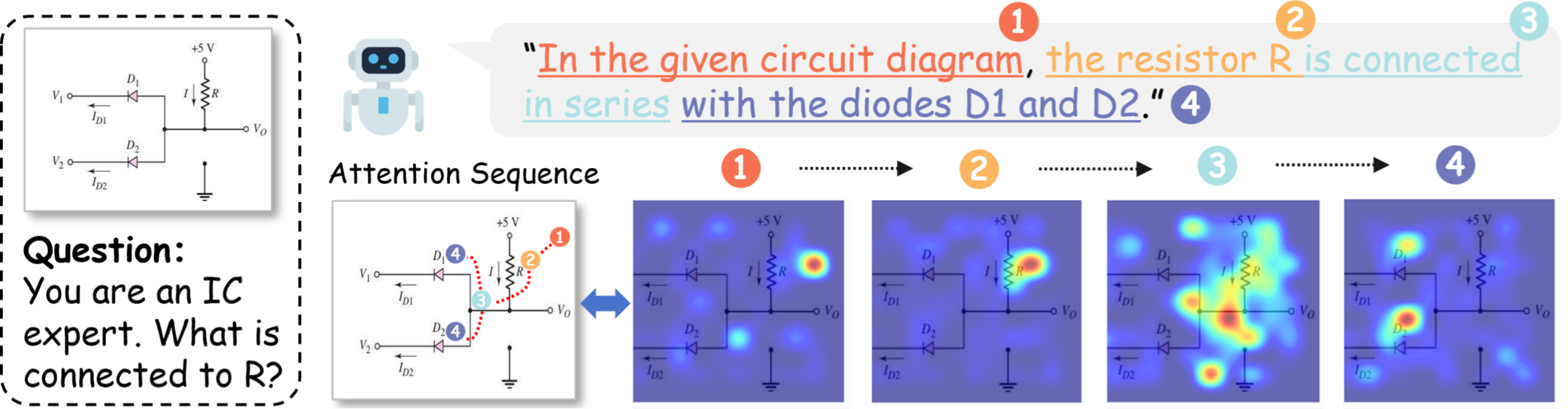} 
  \caption{\textbf{Visualization of MLLM intrinsic attention flow in circuit topology analysis.} The sequence illustrates native visual tracing from the root component, propagating along topological wire paths to terminal devices.}
  \label{fig:attention_flow_org}
\end{figure*}
To investigate how the model intrinsically engages in a native reasoning process to analyze circuit schematic topology without any explicit prompt guidance, we conducted a visual attribution of the MLLM's attention using the Qwen-2.5-VL 7B model. As depicted in \Cref{fig:attention_flow_org}, we identified two key insights:

\begin{itemize}[leftmargin=*]
    \item \textbf{Insight 1: Topology-Guided Attention Flow.} When tackling circuit topology tasks, the MLLM exhibits a visual tracking pattern that mirrors circuit connection logic. Rather than scanning components globally, the model initially anchors on a root component, subsequently traces strictly along topological wire paths, and ultimately shifts focus to downstream devices.
    \item \textbf{Insight 2: Concurrent Branch Tracking.} The MLLM demonstrates the capacity to simultaneously attend to multiple wire branches originating from a single node—a behavior likely facilitated by the multi-head attention mechanism. This parallel processing capability stands in contrast to human visual cognition, which typically favors serial, branch-by-branch analysis.
\end{itemize}
\section{Methods}
In this section, we introduce our Circuit-MLLM framework for sequentially ordered topological reasoning within the latent space. First, we present the latent-space circuit knowledge mining mechanism (\Cref{sec:Latent-Space Circuit Knowledge Mining Mechanism}) to internalize multi-expert visual capabilities without invoking external tools. Next, \Cref{sec:Topology-Guided sequencing strategy} details the topology-guided sequencing strategy, which strictly aligns the model's latent reasoning sequence with rigorous circuit logic. Finally, \Cref{sec:Text-Latent Joint Supervision Training} introduces the text-latent joint supervision training to ensure alignment between internal implicit topological reasoning and explicit textual outputs.
\label{sec:method}
\begin{figure*}[t]
  \centering
  \includegraphics[width=\textwidth]{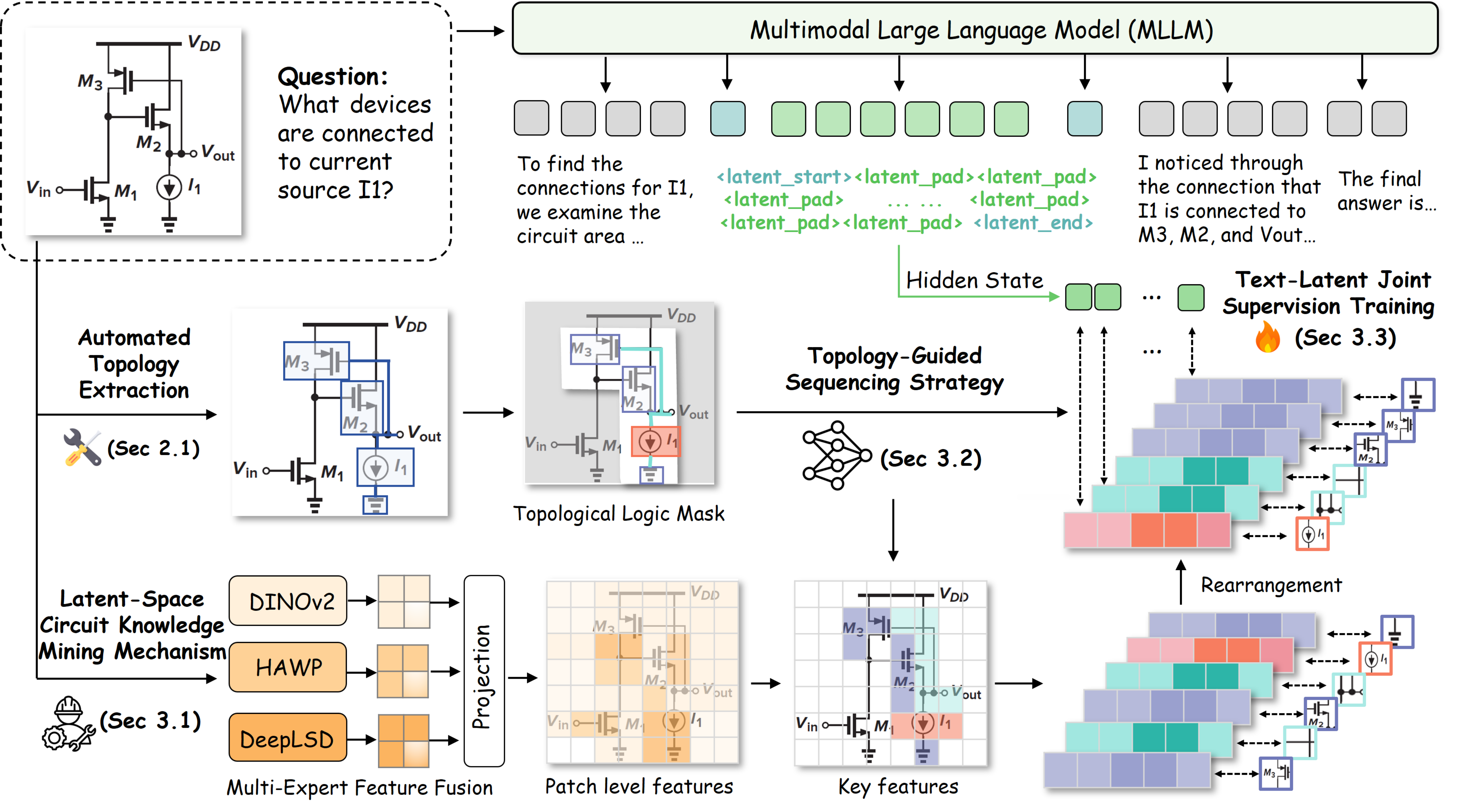} 
  \caption{\textbf{Framework of Circuit-MLLM.} (1) The circuit knowledge mining mechanism based on latent space provides the model with expert-level circuit representation vectors; (2) a Topology-Guided Sequencing Strategy directs latent-space reasoning to strictly follow the circuit's topological order; and (3) Text-Latent Joint Supervision Training unifies latent-space feature alignment and explicit text supervision.}
  \label{fig:overall}
\end{figure*}

\subsection{Latent-Space Circuit Knowledge Mining Mechanism}
\label{sec:Latent-Space Circuit Knowledge Mining Mechanism}

While recent latent-space visual reasoning works show promising results, they primarily rely on generic visual encoders to extract basic image features. However, in specialized domains like circuit analysis, fine-grained visual comprehension (e.g., component topologies) is paramount. To acquire these details, conventional methods often integrate supplementary network modules or explicitly invoke external vision tools during inference. This disjointed approach easily introduces cascading errors and increases inference latency.

To overcome these limitations, we propose a Latent-Space Circuit Knowledge Mining Mechanism. Our core philosophy is to abandon inference-time reliance on external tools, enabling the Circuit-MLLM to intrinsically learn and assimilate multi-expert visual features directly within its latent space during training. To this end, we design a Multi-Expert Feature Fusion module that serves as the source of knowledge guidance.
For a given circuit image $I$, we deploy three heterogeneous vision experts to extract complementary structural and semantic priors: HAWP for holistic wireframe and junction detection ($\mathcal{F}_{\text{HAWP}}$), DeepLSD for fine-grained topological line segment extraction ($\mathcal{F}_{\text{LSD}}$), and DINOv2 for robust patch-level semantic representations ($\mathcal{F}_{\text{DINO}}$).

Due to architectural disparities among these experts, the extracted feature maps vary in spatial resolutions and channel dimensions. To construct a unified representation, we align them to a target spatial grid $(H_s, W_s)$ determined by the Circuit-MLLM's visual encoder. Let $\Phi_{\text{align}}(\cdot)$ denote the bilinear interpolation operator mapping a tensor to this target resolution. The aligned features are concatenated along the channel dimension to form a dense expert representation $\mathcal{F}_{\text{ext}}$:
\begin{equation}
\mathcal{F}_{\text{ext}} = \left[ \Phi_{\text{align}}(\mathcal{F}_{\text{HAWP}}), \Phi_{\text{align}}(\mathcal{F}_{\text{LSD}}), \Phi_{\text{align}}(\mathcal{F}_{\text{DINO}}) \right] \in \mathbb{R}^{(C_1+C_2+C_3) \times H_s \times W_s}.
\end{equation}

Next, we flatten the spatial dimensions and employ a learnable projection matrix $\mathbf{W}_{\text{proj}}$ to project the concatenated heterogeneous features into the identical latent dimension $d$ of the Circuit-MLLM. Let $\mathbf{E}_{\text{orig}} \in \mathbb{R}^{N \times d}$ denote the initial visual tokens extracted by the Circuit-MLLM's original visual encoder, where $N = H_s \times W_s$. To imbue these initial tokens with circuit-specific priors, we concatenate them and pass them through a fusion layer $\mathbf{W}_{\text{fuse}}$ to obtain the knowledge-enriched feature pool $\mathbf{E}_{\text{fused}}$:
\begin{align}
\mathbf{E}_{\text{expert}} &= \text{Flatten}(\mathcal{F}_{\text{ext}}) \mathbf{W}_{\text{proj}}, \\
\mathbf{E}_{\text{fused}} &= \left[ \mathbf{E}_{\text{orig}}, \mathbf{E}_{\text{expert}} \right] \mathbf{W}_{\text{fuse}}.
\end{align}
The fused feature pool $\mathbf{E}_{\text{fused}}$ provides a set of candidate alignment targets for \Cref{sec:Topology-Guided sequencing strategy} and \Cref{sec:Text-Latent Joint Supervision Training}, forcing the MLLM to internalize expert-enhanced visual semantics within its latent space.

\subsection{Topology-Guided Sequencing Strategy}
\label{sec:Topology-Guided sequencing strategy}
Motivated by the unique attention behaviors observed in our visual attribution analysis in \Cref{sec:Visual Attribution}, we introduce the \textbf{Topology-Guided Sequencing Strategy}. This latent-space reasoning paradigm explicitly guides the model to follow the sequential perspective of topological logic, aligning its inherent tracking capabilities with the rigorous demands of circuit analysis.

Instead of relying on coarse bounding boxes that introduce background noise, we utilize the automated topology extraction workflow (\Cref{sec:Topology Extraction}) to obtain exact pixel-level localization for each topological query. Guided by the visual attention shifts of Circuit-MLLM and human domain expertise, we formulate a pixel-wise sequencing map, termed the \textbf{Topological Logic Mask} $S(p)$:

\begin{equation}
S(p) = \begin{cases}
0, & \text{if } p \in \Omega_{\text{bg}} \\ 
1, & \text{if } p \in \Omega_{\text{root}} \\ 
2 + \frac{\mathcal{D}_{\text{BFS}}(p, \Omega_{\text{root}})}{\max_{q \in \Omega_{\text{wire}}} \mathcal{D}_{\text{BFS}}(q, \Omega_{\text{root}})}, & \text{if } p \in \Omega_{\text{wire}} \\ 
3 + \gamma_i, & \text{if } p \in \Omega_{\text{conn}}^{(i)}
\end{cases}
\end{equation}
where $\mathcal{D}_{\text{BFS}}$ denotes the topological distance calculated via Breadth-First Search, capturing the "near-to-far" principle of visual attention. Meanwhile,$\gamma_i \sim \mathcal{U}(0, 1)$ assigns a distinct continuous value to differentiate connected devices. 

To align this pixel-level mask with the patch-level features $\mathbf{E}_{\text{fused}}$, we apply Adaptive Max Pooling, yielding the patch-level Topological Logic Mask $\hat{S}_{i,j} = \max_{p \in \mathcal{P}_{i,j}} S(p)$. This acts as a conservative filter to preserve highly localized structures like thin wires. 
Crucially, after filtering out background features ($\hat{S}_{i,j} > 0$), we explicitly inject topological logic by defining a permutation $\pi$ on the valid feature indices $\mathcal{V}$. This permutation sorts the indices such that their corresponding mask values are monotonically increasing: $\hat{S}_{\pi(1)} \le \hat{S}_{\pi(2)} \le \dots \le \hat{S}_{\pi(|\mathcal{V}|)}$. This seamlessly transforms the unordered 2D spatial features into a strictly topologically-ordered 1D sequence:

\begin{equation}
\tilde{\mathbf{E}}_{\text{fused}} = [\mathbf{E}_{\text{fused},\pi(1)}, \mathbf{E}_{\text{fused},\pi(2)}, \dots, \mathbf{E}_{\text{fused},\pi(|\mathcal{V}|)}].
\end{equation}

To accommodate the MLLM's fixed latent capacity $k$, a 1D Adaptive Average Pooling compresses this sequence into a fixed-length target $\mathbf{E}^* \in \mathbb{R}^{k \times d}$:

\begin{equation}
\mathbf{E}^*_m = \frac{1}{|B_m|} \sum_{t \in B_m} \tilde{\mathbf{E}}_{\text{fused},t}, \quad m \in \{1, 2, \dots, k\}.
\end{equation}
This ordered, expert-enriched sequence serves as the target latent vectors for knowledge guidance. During the autoregressive generation process, we employ a \textbf{Sequential Topological-Knowledge Alignment Loss} ($\mathcal{L}_{\text{topo}}$) to optimize the model by minimizing the cosine distance between the hidden state $h_m$ at the final layer and the target latent vectors:

\begin{equation}
\mathcal{L}_{\text{topo}} = \frac{1}{k} \sum_{m=1}^{k} \left( 1 - \cos(h_m, \mathbf{E}^*_m) \right).
\end{equation}
This mechanism effectively constrains the model to traverse predefined topological trajectories, thereby anchoring its latent representations within a logically structured and dynamically sequenced reasoning space.

\subsection{Text-Latent Joint Supervision Training}
\label{sec:Text-Latent Joint Supervision Training}
While the latent tokens are supervised by the unified topological knowledge alignment objective $\mathcal{L}_{\text{topo}}$, the surrounding text tokens are optimized via a standard autoregressive cross-entropy loss. Let $\boldsymbol{x}$ denote the input query and circuit image, and $\boldsymbol{o}_{\text{pre}}$ / $\boldsymbol{o}_{\text{post}}$ denote the text segments respectively preceding and succeeding the generated latent tokens $\{h_m\}_{m=1}^k$. The textual loss $\mathcal{L}_{\text{text}}$ is formulated as:
\begin{equation}
\begin{split}
\mathcal{L}_{\text{text}} &= \sum_{i=1}^{|\boldsymbol{o}_{\text{pre}}|} \ell_{\text{CE}}\left(\boldsymbol{o}_{\text{pre},i}, P_\theta(\boldsymbol{x}, \boldsymbol{o}_{\text{pre},<i})\right) \\
&\quad + \sum_{i=1}^{|\boldsymbol{o}_{\text{post}}|} \ell_{\text{CE}}\left(\boldsymbol{o}_{\text{post},i}, P_\theta(\boldsymbol{x}, \boldsymbol{o}_{\text{pre}}, \{h_m\}_{m=1}^k, \boldsymbol{o}_{\text{post},<i})\right),
\end{split}
\end{equation}
where $P_\theta(\cdot)$ denotes the next-token prediction probability of the Circuit-MLLM. The overall training objective seamlessly combines both terms:
\begin{equation}
\mathcal{L}_{\text{total}} =  \lambda \mathcal{L}_{\text{topo}} +\mathcal{L}_{\text{text}},
\end{equation}
where $\lambda$ is a balancing coefficient. This joint optimization anchors the latent tokens in the topologically-ordered expert visual space while weaving them naturally into the model's explicit textual reasoning.

\section{Experiments}
\label{sec:Experiments}

\subsection{Experimental Settings}
\label{sec:Experimental Settings}
\textbf{Benchmarks}. We evaluate our method on Circuit-MLLM-bench, a custom large-scale circuit benchmark. This benchmark is constructed via the Automated Circuit Schematic Topology Extraction pipeline detailed in \Cref{sec:Topology Extraction} and has been rigorously verified by human experts. The benchmark evaluates models across two primary categories: topological analysis tasks, consisting of Connection Judgment and Connection Identification; and conventional detection tasks, which include Total Count, Type Count, and Element Class. Detailed construction procedures are provided in the supplementary material.

\noindent\textbf{Data Synthesis}. Driven by our automated topology extraction workflow, we generate question-answer (QA) pairs for each circuit image across the five tasks, alongside auxiliary images and pixel-level masks that delineate semantic regions and govern latent topological sequencing. We sample 5,000 instruction-tuning instances distributed by task complexity: Connection Identification (30\%), Connection Judgment (25\%), Element Class (15\%), Total Count (15\%), and Type Count (15\%). To prevent data leakage, training images are strictly mutually exclusive from the Circuit-MLLM-bench evaluation set.

\noindent\textbf{Baselines}. We compare our approach against four baseline categories: (1) General-Purpose MLLMs: Mainstream models, encompassing closed-source models (e.g., GPT 5.1,GPT-4o, Claude-4-Sonnet, Doubao-1.5-vision-pro) and open-source models (e.g., Qwen 2.5 and Qwen 3 series, DeepSeek-VL2, and GLM-4.5V). (2) Latent-Space Visual Reasoning: Mirage \cite{yang2025machine} and ILVR\cite{dong2025interleaved}.(3) Tool-Augmented Circuit MLLMs: Masala CHAI \cite{bhandari2024masala}, representing methods that explicitly invoke external tools to modify images prior to analysis. (4) Direct Fine-Tuning: Supervised Fine-Tuning (SFT) and SFT combined with Group Relative Policy Optimization (SFT+GRPO), representing models trained directly on our dataset without utilizing the latent-space mechanism.

\noindent\textbf{Implementation Details}. All experiments employ Qwen2.5-VL 7B as the foundational base model. Our framework is implemented using PyTorch and trained on 8 NVIDIA A100 GPUs. We conduct supervised fine-tuning for 15 epochs, utilizing a batch size of 8 alongside a cosine learning rate scheduler with an initial learning rate of 1e-5 across both stages. The latent token size is set to $k = 4$, and the loss balancing coefficient is configured to $\gamma = 0.6$.
\subsection{Experimental Results}
\label{sec:Experimental Results}

\begin{table*}[t]
\caption{\textbf{Performance comparison across all task categories.} ``Acc.'', ``F1'', and ``EMR'' denote Accuracy, F1-score, and Exact Match Ratio, respectively. ``Acc.'' is evaluated on single-choice and counting tasks, whereas ``F1'' and``EMR" are applied to multiple-response tasks. ``Avg.'' represents the overall average score.}
\label{tab:Circuit-MLLM-bench}
\centering
\setlength{\tabcolsep}{1.pt}
\resizebox{\textwidth}{!}{
    \begin{tabular}{llcccccccc}
    \toprule
    \multicolumn{1}{l}{\multirow{3}{*}{\textbf{Models}}} & \multicolumn{1}{l}{\multirow{3}{*}{\textbf{Size}}} & \multicolumn{6}{c}{\textbf{Circuit-MLLM-bench}} \\
    \cmidrule(lr){3-9}
     & & 
    \multicolumn{1}{c}{Total Count} & \multicolumn{1}{c}{Type Count} & \multicolumn{1}{c}{Element Class} & \multicolumn{1}{c}{Conn.Judge} & \multicolumn{2}{c}{Conn.Identi} & \multirow{2}{*}{Avg}\\
    \cmidrule(lr){3-3} \cmidrule(lr){4-4} \cmidrule(lr){5-5} \cmidrule(lr){6-6}  \cmidrule(lr){7-8}
     & & 
    Acc.$\uparrow$ & Acc.$\uparrow$ & Acc.$\uparrow$ & Acc.$\uparrow$ & F1$\uparrow$ & EMR.$\uparrow$\\
    \midrule
    GPT 5.1 & /  &43.60  &	60.62  &	96.49  &	61.11 & 	76.08  &	30.20  &	61.35\\ 
    GPT-4o & / & 42.01 &60.08  &97.40  &58.50  &78.68    &36.30  &62.16\\
    Claude-4-Sonnet &/ &44.91  &58.23  &95.13  &57.13  &69.54   & 24.80& 58.17\\ 
    Doubao-1.5-vision-pro &/ &40.21  &65.00  &91.35  & 60.36 &74.08  &24.40 & 59.23\\
    deepseek-vl2 &27B &24.13  &43.80  &77.62  &56.56  & 59.61  & 17.00 &46.45\\
    GLM-4.5V &106B &58.62  &59.50   &90.08   &58.59  & 79.37 & 40.00&64.36   \\
    \midrule
    \multirow{3}{*}{Qwen2.5-VL} 
    & 7B  &21.06  &48.40  &83.55  &55.92  &65.53  &15.20  &48.28\\
    & 32B &26.77  & 57.60  & 88.02  & 60.14 &68.77   &14.80   &52.68 \\
    & 72B &29.95  &58.10   &90.60  &57.31  &72.73  &24.60  &55.55\\ 
    \midrule
    \multirow{2}{*}{Qwen3-VL}
    & 8B  &39.47   & 56.00 &90.30   &57.52  &76.86   &38.30 &59.74 \\
    & 32B &49.63  & 60.00 &92.95  &58.59  &82.02 & 42.90 &64.35\\
    \midrule
    Masala-CHAI & 7B &27.72  &48.10  &93.10  & 51.00 &66.35&19.10  & 50.88 \\
    \textbf{Circuit-MLLM} & 7B  &\textbf{75.87}  &\textbf{65.10} &\textbf{97.40}  &\textbf{72.60}  &\textbf{88.90} & \textbf{57.30} &\textbf{76.20}\\
    \bottomrule
    \end{tabular}    
}
\end{table*}
\textbf{Results on all task categories.} \Cref{tab:Circuit-MLLM-bench} presents the comparative evaluation results of various open-source and closed-source models across all task categories on Circuit-MLLM-Bench. In terms of average performance, Circuit-MLLM outperforms all existing baseline models, achieving a significant 57.8\% relative improvement over its backbone, Qwen-2.5-VL-7B, and an approximate 22.5\% improvement over the leading model GPT-4o. This underscores the effectiveness of Circuit-MLLM in solving complex circuit-related problems. Furthermore, although Masala-CHAI, a method utilizing external visual tools, holds a slight edge over Qwen-2.5-VL-7B in total count, element class, and connection identification, its overall gain is merely 4\%, indicating the limitations of relying purely on external tools. Most importantly, on the highly challenging connection identification task, our method achieves a breakthrough in the Exact Match Ratio (EMR), reaching approximately three times that of the backbone model, thereby demonstrating its robust capability in parsing complex topologies.

\paragraph{\textbf{\textup{Results on topological analysis tasks.}}}To evaluate Circuit-MLLM in topological analysis, we fine-tuned the latent-space visual baselines, Mirage \cite{yang2025machine} and ILVR \cite{dong2025interleaved}, using our latent-space dataset with the same latent size of 4 to ensure a fair comparison. As shown in \Cref{tab:Results on topological analysis tasks.}, Circuit-MLLM consistently outperforms both Mirage and ILVR across all difficulty levels in Connection Judgment and Connection Identification. The average score increases by approximately 5\%, demonstrating the overall effectiveness of our method. Notably, on the Hard difficulty of Connection Identification, our EMR exhibits a substantial 25\% improvement over ILVR. Furthermore, Masala-CHAI performs almost identically to the backbone model, Qwen-2.5-VL-7B, on topological tasks, suggesting that explicitly modifying the original image yields marginal benefits. Compared to the SFT+GRPO baseline, Circuit-MLLM achieves an overall improvement of about 7\%, with maximum gains reaching 13\% on specific tasks. Finally, latent-space visual methods generally surpass the direct SFT baseline (without latent mechanisms) across topological tasks. This consistent superiority strongly proves the validity of conducting topological analysis directly within the latent space.
\begin{table*}[t]
\caption{\textbf{Performance comparison on topological analysis tasks.} The evaluation encompasses two tasks: Connection Judgment and Connection Identification. The difficulty levels (Easy, Medium, and Hard) are determined by the number of connected components and the total number of components within the circuit diagram.}
\centering
\label{tab:Results on topological analysis tasks.}
\small
\setlength{\tabcolsep}{3.pt}
\resizebox{\textwidth}{!}{
    \begin{tabular}{ccccccccccccc}
    \toprule
    \multicolumn{1}{c}{\multirow{3}{*}{\textbf{Models}}} & \multicolumn{2}{c}{\multirow{3}{*}{\textbf{Methods}}} & \multicolumn{3}{c}{Connection Judge.} & \multicolumn{6}{c}{Connection Identification} & \multirow{3}{*}{Avg}\\
    \cmidrule(lr){4-6} \cmidrule(lr){7-12}
    & & & \multicolumn{1}{c}{Easy} & \multicolumn{1}{c}{Medium} & \multicolumn{1}{c}{Hard} & \multicolumn{2}{c}{Easy} & \multicolumn{2}{c}{Medium} & \multicolumn{2}{c}{Hard} & \\
    \cmidrule(lr){4-4} \cmidrule(lr){5-5} \cmidrule(lr){6-6} \cmidrule(lr){7-8} \cmidrule(lr){9-10} \cmidrule(lr){11-12}
    & & & Acc.$\uparrow$ & Acc.$\uparrow$ & Acc.$\uparrow$ & F1$\uparrow$ & EMR.$\uparrow$ & F1$\uparrow$ & EMR.$\uparrow$ & F1$\uparrow$ & EMR.$\uparrow$ & \\
    \midrule
    \multirow{5}{*}{\begin{tabular}[c]{@{}c@{}}Qwen2.5-VL\\7B\end{tabular}}
    & SFT & GRPO & & & & & & & & & & \\
    \cmidrule(lr){2-13}
    &&  &59.29 	&51.56 &	41.52 &	75.24 &	44.01 &	61.70 &	7.20 	&61.54 	&3.78 	&45.09 \\
    & \checkmark &  & 76.70 & 63.28 & 56.68 &87.86 	&69.46 	&89.92&59.20&80.40 &26.12 &	67.74  \\
    & \checkmark & \checkmark&76.70 &	62.76 &	57.40 	&87.86 	&69.76 &	89.85 &	58.13 	&80.53 	&26.80& 	67.75  \\
    \cmidrule(lr){2-13}
    & \multicolumn{2}{c}{Masala-CHAI}&55.16 &	53.91& 	41.88 	&75.70 &	45.21 &	61.87 	&6.93 	&61.46 &	4.47 	&45.18 \\
    \midrule
    \multirow{3}{*}{\begin{tabular}[c]{@{}c@{}}Latent\\Reasoning\\7B\end{tabular}} 
    & \multicolumn{2}{c}{Mirage}  &77.88 &	66.15 &	64.98 &	90.41 &	72.16 	&89.48& 	58.67 &	81.88 &	26.80& 	69.82  \\
    & \multicolumn{2}{c}{ILVR}    &79.06 	&63.28 	&63.90 	&89.74 	&72.46 	&88.98 	&55.73 &	79.47 &	22.68 	&68.37 \\
    & \multicolumn{2}{c}{\textbf{Circuit-MLLM}} & \textbf{81.42} & \textbf{70.31} & \textbf{64.98} & \textbf{91.44} & \textbf{76.35} & \textbf{91.10} & \textbf{62.67} & \textbf{83.54} & \textbf{28.52} & \textbf{72.26} \\
    \bottomrule
    \end{tabular}    
}
\end{table*}

\subsection{Ablation Study}
\label{sec:Ablation Study}
\paragraph{\textbf{\textup{Effectiveness of the Proposed Method.}}}
\begin{table*}[t]
\caption{\textbf{Ablation Study of our methods.} ``Seq.'' and ``Expert'' refer to our two core components: the Topology-Guided Sequencing Strategy and Latent-Space Circuit Knowledge Distillation, respectively.}
\centering
\label{tab:ablation_study}
\small
\setlength{\tabcolsep}{3.pt}
\resizebox{\textwidth}{!}{
    \begin{tabular}{ccccccccccccc}
    \toprule
    \multicolumn{1}{c}{\multirow{3}{*}{\textbf{Models}}} & \multicolumn{2}{c}{\textbf{Method}} & \multicolumn{3}{c}{Connection Judge.} & \multicolumn{6}{c}{Connection Identification} & \multirow{3}{*}{Avg}\\
    \cmidrule(lr){2-3} \cmidrule(lr){4-6} \cmidrule(lr){7-12}
    & \multirow{2}{*}{\textbf{Seq.}} & \multirow{2}{*}{\textbf{Expert}} & \multicolumn{1}{c}{Easy} & \multicolumn{1}{c}{Medium} & \multicolumn{1}{c}{Hard} & \multicolumn{2}{c}{Easy} & \multicolumn{2}{c}{Medium} & \multicolumn{2}{c}{Hard} & \\
    \cmidrule(lr){4-4} \cmidrule(lr){5-5} \cmidrule(lr){6-6} \cmidrule(lr){7-8} \cmidrule(lr){9-10} \cmidrule(lr){11-12}
    & & & Acc.$\uparrow$ & Acc.$\uparrow$ & Acc.$\uparrow$ & F1$\uparrow$ & EMR.$\uparrow$ & F1$\uparrow$ & EMR.$\uparrow$ & F1$\uparrow$ & EMR.$\uparrow$ & \\
    \midrule
    \multirow{4}{*}{\begin{tabular}[c]{@{}c@{}}Latent\\Reasoning\\7B\end{tabular}} 
    & &  &77.88 	&65.10 	&62.45 	&89.97 	&73.05 	&90.60 	&61.87 &82.28& 	25.46 &	69.85  \\
    & \checkmark &   &79.06 	&66.15 	&64.26 	&89.99 &73.05 	&90.98& 	62.40& 	\textbf{83.86} &	25.77 	&70.61 \\
    & & \checkmark   & 77.88 & 65.62 & 62.50 & 90.64 & 73.35 & 	\textbf{91.22} & \textbf{63.47} & 83.83&  27.15&  70.64 \\
    & \checkmark & \checkmark & \textbf{81.42} & \textbf{70.31} & \textbf{64.98} & \textbf{91.44} & \textbf{76.35} & 91.10 & 62.67 & 83.54 & \textbf{28.52} & \textbf{72.26} \\
    \bottomrule
    \end{tabular}    
}
\end{table*}
As shown in \Cref{tab:ablation_study}, the baseline latent reasoning model achieves an average score of 69.85. Introducing the Topology-Guided Sequencing Strategy alone improves the average score to 70.61, bringing consistent gains in Connection Judgment across all difficulty levels. Conversely, incorporating only the Latent-Space Circuit Knowledge mining mechanism raises the average score to 70.64, showing particular effectiveness in the Connection Identification task. When both components are integrated, Circuit-MLLM achieves the highest overall average score of 72.26, a 2.41 improvement over the baseline. The full model demonstrates superior performance across all Connection Judgment tasks and attains the highest EMR (28.52) on the Hard difficulty of Connection Identification. These quantitative results confirm that the two components are complementary, effectively enhancing the model's topological analysis capabilities within the latent space.

\begin{table*}[t]
\centering
\caption{\textbf{Ablation Study of Latent Size.} The latent size denotes the total number of latent vectors generated during the inference phase. Across all experiments in this section, the loss balancing coefficient is uniformly configured to $\gamma = 0.6$.}
\label{tab:Ablation Study of latent size}
\setlength{\tabcolsep}{3.pt}
\resizebox{\textwidth}{!}{
    \begin{tabular}{ccccccccc}
    \toprule
    \multicolumn{1}{l}{\multirow{2}{*}{\textbf{Models}}} & \multicolumn{1}{l}{\multirow{2}{*}{\textbf{Latent Size}}} & \multicolumn{1}{c}{Total Count} & \multicolumn{1}{c}{Type Count} & \multicolumn{1}{c}{Element Class} & \multicolumn{1}{c}{Conn.Judge} & \multicolumn{2}{c}{Conn. Ident} & \multirow{2}{*}{Avg}\\
    \cmidrule(lr){3-3} \cmidrule(lr){4-4} \cmidrule(lr){5-5} \cmidrule(lr){6-6}  \cmidrule(lr){7-8}
     & &  Acc.$\uparrow$ & Acc.$\uparrow$ & Acc.$\uparrow$ & Acc.$\uparrow$ & F1$\uparrow$ & EMR.$\uparrow$ \\
    \midrule
    \multirow{4}{*}{\begin{tabular}[c]{@{}c@{}}Latent\\Reasoning\\7B\end{tabular}}
    &2 &74.05&66.60 &96.20&67.70&87.52&54.50  &74.42\\
    &4 &\textbf{75.87}  &65.10  &\textbf{97.40}  &\textbf{72.60} & 88.92&\textbf{57.30} &\textbf{76.20}\\
    &6 &73.75&66.70 &95.10&67.80&87.49&54.30  &74.19\\
    &8 &75.13&\textbf{67.80} &96.50&67.70&\textbf{89.12} &56.30  &75.43 \\
    \bottomrule
    \end{tabular}    
}
\end{table*}

\begin{table*}[t]
\centering
\caption{\textbf{Ablation study of circuit experts.} We evaluate the impact of different expert models, including DINOv2 , HAWP (Holistically-Attracted Wireframe Parsing, and DeepLSD (Deep Line Segment Detection). In all configurations, the latent size is set to 6, and the loss balancing coefficient is uniformly configured to $\gamma = 0.6$.}
\label{tab:Ablation_study_of_circuit_experts}
\setlength{\tabcolsep}{3.pt}
\resizebox{\textwidth}{!}{
    \begin{tabular}{cccccccccc}
    \toprule
    \multicolumn{3}{c}{\textbf{Circuit Experts}} & \multicolumn{1}{c}{Total Count} & \multicolumn{1}{c}{Type Count} & \multicolumn{1}{c}{Element Class} & \multicolumn{1}{c}{Conn.Judge} & \multicolumn{2}{c}{Conn. Ident} & \multirow{2}{*}{Avg}\\
    \cmidrule(lr){1-3} \cmidrule(lr){4-4} \cmidrule(lr){5-5} \cmidrule(lr){6-6} \cmidrule(lr){7-7} \cmidrule(lr){8-9}
    DINO & HAWP & DeepLSD & Acc.$\uparrow$ & Acc.$\uparrow$ & Acc.$\uparrow$ & Acc.$\uparrow$ & F1$\uparrow$ & EMR.$\uparrow$ & \\
    \midrule
    \checkmark &  &            & 75.23 &\textbf{66.80}  &94.50 &69.00 &88.60 &56.80  &75.16 \\
     &\checkmark            &            &71.11 &65.90 &95.70 &70.60 &\textbf{89.16} &57.20 &74.95\\
    \checkmark       &\checkmark       & &74.49 &66.50 &96.40 &71.00 &88.16 &54.80 &75.23 \\
    \checkmark & \checkmark & \checkmark &\textbf{75.87}  &65.10  &\textbf{97.40}  &\textbf{72.60} & 88.92&\textbf{57.30} &\textbf{76.20} \\
    \bottomrule
    \end{tabular}    
}
\end{table*}

\paragraph{\textbf{\textup{Selection of latent size.}}}
\Cref{tab:Ablation Study of latent size} presents the performance of Circuit-MLLM on the Circuit-MLLM-Bench across different latent sizes. In terms of the overall score, a latent size of $4$ yields the best results, representing the most balanced choice. Furthermore, this indicates that $4$ latent pads are sufficient to encapsulate the reasoning process for conventional circuit problems.

\paragraph{\textbf{\textup{Effectiveness of different circuit visual experts.}}}
\Cref{tab:Ablation_study_of_circuit_experts} presents the performance of our three circuit visual experts, DINOv2, HAWP, and DeepLSD, across five circuit tasks. Employing all experts simultaneously yields the highest average score, demonstrating their collective effectiveness in circuit problem analysis. Furthermore, the individual contributions of each expert vary significantly across different problem types. For instance, utilizing only DINOv2 leads to strong performance on counting tasks, such as Total Count and Type Count, highlighting its precision in general object recognition; however, it struggles with connection-related problems. Conversely, relying solely on HAWP yields better results on Conn. Judge and Conn. Ident., verifying that visual experts specialized in wire detection are indeed highly effective for topology-related tasks.

\paragraph{\textbf{\textup{Additional Results.}}} Further experimental results, including coefficient $\gamma$ selection and Amsbench \cite{shi2025amsbench} evaluations, are provided in the supplementary material.
\section{Analysis}
\label{sec:Analysis}

\begin{figure*}[t]
  \centering
  \includegraphics[width=0.95\textwidth]{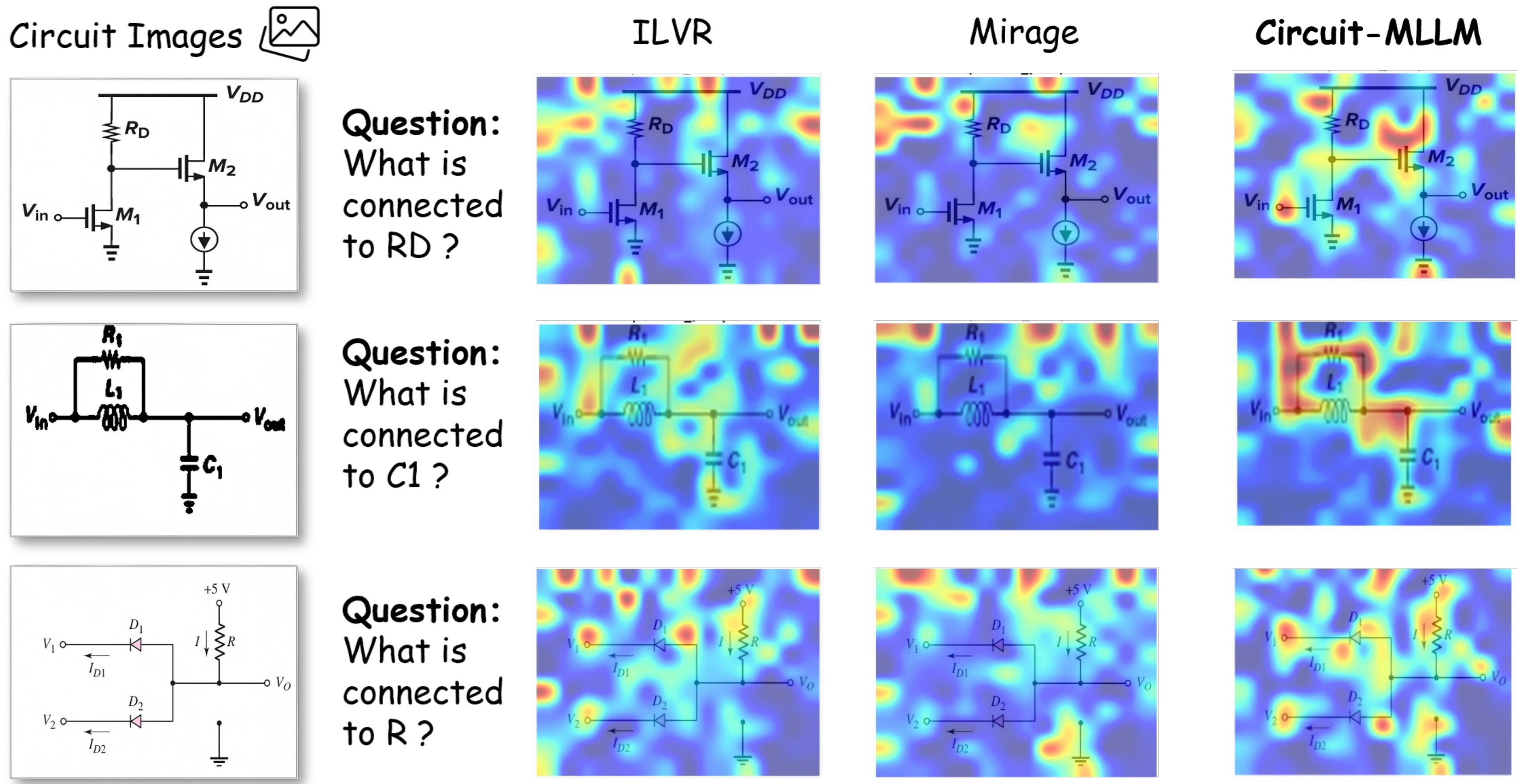} 
   \caption{\textbf{Visual comparisons in the latent space.} The latent features are captured prior to the generation of each <latent pad> token. Compared to Mirage and ILVR, our Circuit-MLLM achieves both higher fine-grained resolution and precise localization.}
  \label{fig:latent_visual_compare}
\end{figure*}
\paragraph{\textbf{\textup{Latent Space Attention Analysis.}}}To investigate the visual regions focused on by the MLLM in the latent space, we extract the latent features from the last layer prior to the generation of each <latent pad> token. These features are utilized as latent vectors and mapped to the input image to generate heatmaps as shown in \Cref{fig:latent_visual_compare}. Observations indicate that when the input text queries specific components, our Circuit-MLLM accurately focuses on the root component, adjacent wires, and the final target component within the latent space. In contrast, Mirage and ILVR struggle to localize the regions of interest and exhibit lower fine-grained resolution, making it difficult to capture minute details such as wires. This demonstrates that our proposed Topological Logic Mask and Visual Expert effectively assist the model in capturing these crucial circuit topology features. Furthermore, we note that the latent space naturally avoids noise components and blank areas, proving that the latent space visual approach yields superior performance compared to direct image cropping.

\begin{figure*}[t]
  \centering
  \includegraphics[width=\textwidth]{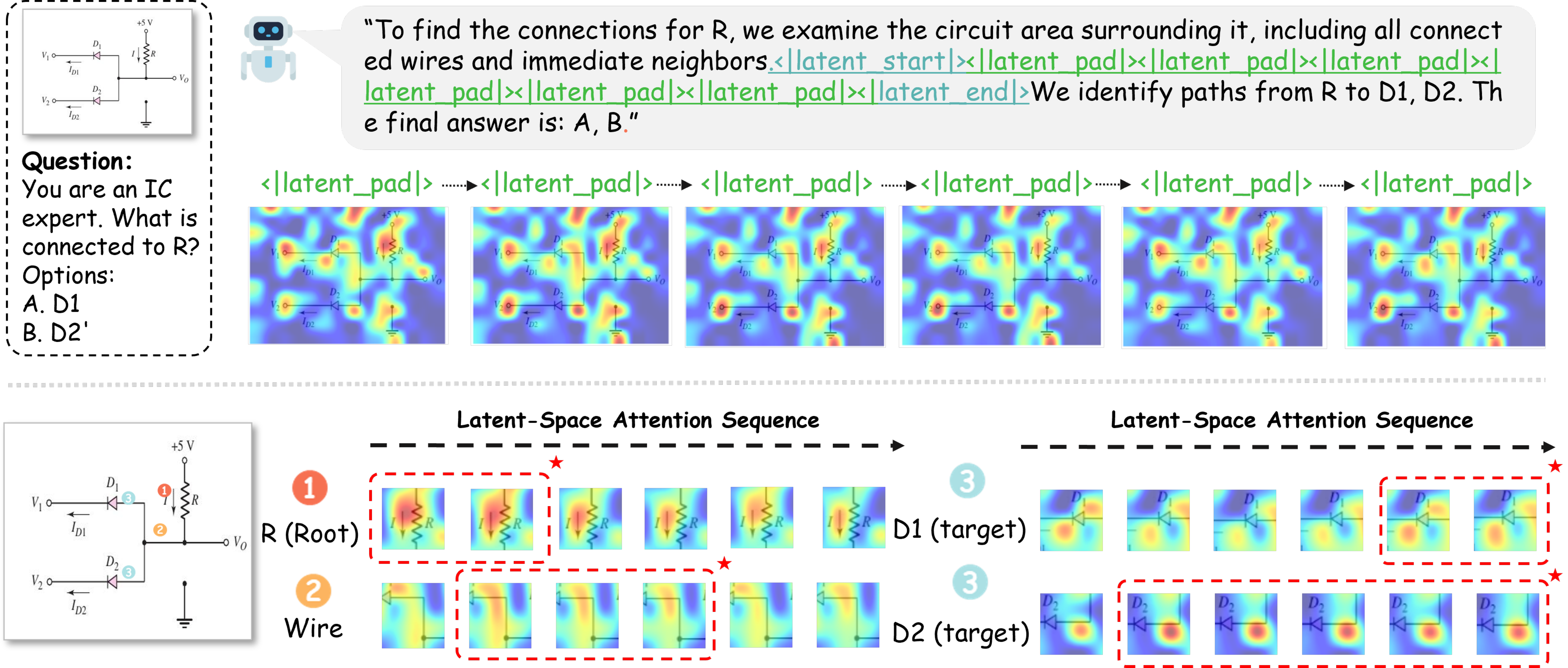} 
    \caption{\textbf{Dynamic visual shifts within the latent space.} We demonstrate this using latent size of 6. \textbf{Top:} Global heatmaps of the penultimate layer's latent representations overlaid on the input image. \textbf{Bottom:} Localized heatmaps focusing on critical regions for topological reasoning, including the root device, wire, and target device.}
  \label{fig:visual_latent}
\end{figure*}
\paragraph{\textbf{\textup{Latent Space Visual Shift Analysis.}}}
To verify whether Circuit-MLLM achieves dynamic visual shifts in the latent space, we extracted six consecutive latent feature vectors as shown in the Top of \Cref{fig:visual_latent}. We specifically focus on the critical regions for topological reasoning: the root device, the wire, and the target device. As illustrated in the Bottom of \Cref{fig:visual_latent}, a temporal progression of visual attention is observed: the model initially focuses heavily on the root device, followed by a gradual decay of attention in this area; the visual focus then shifts along the connecting wires, ultimately concentrating on the target device. This trajectory aligns with human analytical workflow, proving Circuit-MLLM performs accurate topological reasoning entirely within the latent space.
\section{Conclusion}
\label{sec:Conclusion}
In this paper, we proposed Circuit-MLLM, the first topological logic-guided latent-space visual reasoning framework dedicated to complex circuit topology analysis. To address the inherent irregularity of topological regions in circuit diagrams, we pioneered the embedding of the topological reasoning process into the latent space. Extensive experiments demonstrate that by incorporating the circuit knowledge mining mechanism and the topology-guided sequencing strategy, Circuit-MLLM accurately focuses on the intended topological regions, leading to significant performance improvements in complex topological analysis tasks.

\section*{Acknowledgements}
This work was supported by NSFC (Grant No. U25A20485, W2412034).

%
%
\bibliographystyle{splncs04}
\bibliography{main}
\clearpage
\appendix
\section{Additional Experimental Results}
\label{sec:Additional Experimental Results}
\subsection{Chain-of-Thought Paradigms on Topological Tasks}

This section evaluates alternative Chain-of-Thought (CoT) paradigms on topology-related tasks to justify our architectural design. We compare two baselines, Visual CoT and Textual CoT, against our proposed Latent-Visual approach:

\paragraph{\textbf{\textup{Visual CoT.}}} Drawing inspiration from Visual SKETCHPAD and Visual CoT, we decompose the reasoning process into a two-stage pipeline. First, the model identifies the relevant topological regions and triggers visual cropping operations to extract and enlarge these areas. Subsequently, both the original global image and the cropped local patches are fed back into the model to formulate a comprehensive evaluation and yield the final result.

\begin{figure*}[h]
  \centering
  \includegraphics[width=\textwidth]{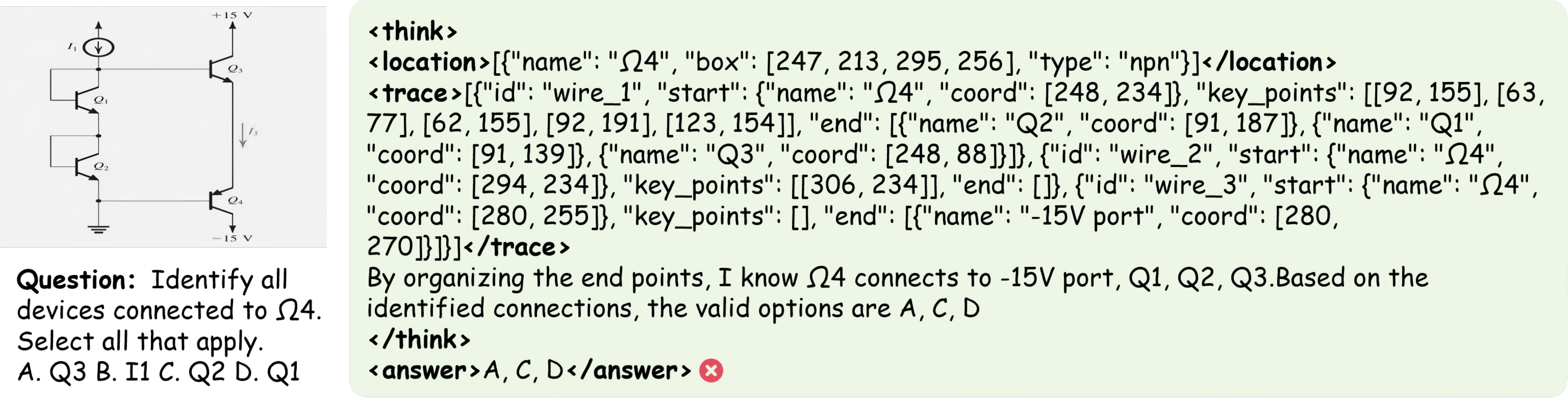} 
  \caption{\textbf{Illustration of the Textual CoT reasoning process for topology analysis, }detailing three sequential stages: locating, tracing, and judging.}
  \label{fig:text_cot}
\end{figure*}

\paragraph{\textbf{\textup{Textual CoT.}}} As illustrated in \Cref{fig:text_cot}, we formulate the determination of connection relationships as a sequential three-step process: locating the root component, tracing along the connecting lines to extract key node coordinates, and ultimately judging the target connected component.

\begin{figure*}[h]
  \centering
  \includegraphics[width=\textwidth]{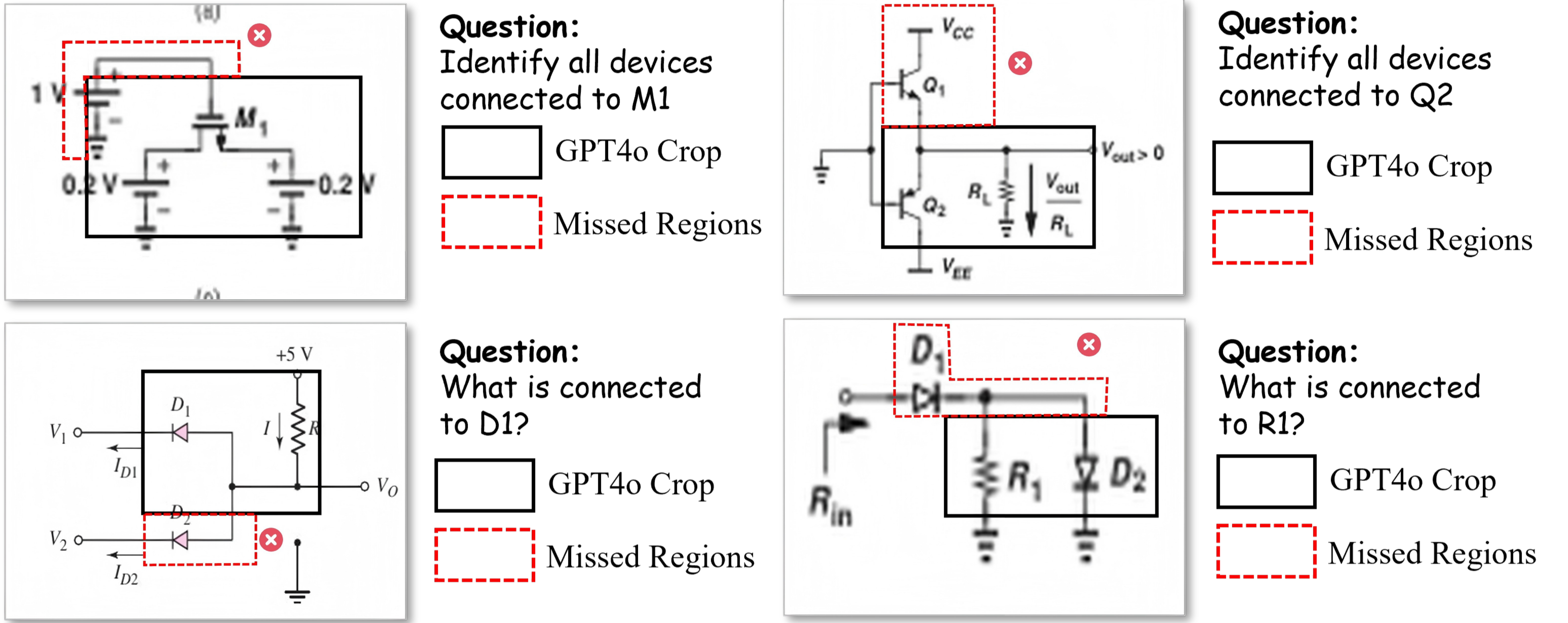} 
  \caption{\textbf{Question-guided topology region cropping by GPT-4o. }The black bounding boxes indicate the corresponding regions successfully cropped by the model based on the given questions, while the red dashed boxes highlight the missed regions.}
  \label{fig:crop}
\end{figure*}

\paragraph{\textbf{\textup{Results.}}} The performance comparison is summarized in Table \ref{tab:Chain-of-Thought}. First, using GPT-4o as the foundation model, we evaluate Visual CoT. While Visual CoT achieves a marginal improvement on the Connection Identification task, it degrades on the Connection Judge task. We attribute this degradation to the difficulty of accurately localizing fine-grained wire regions (as illustrated in \Cref{fig:crop}); the hard-cropping operation inevitably disrupts continuous topological structures and introduces visual artifacts. 
Furthermore, using Qwen2.5-VL 7B, we benchmark our proposed Latent-Visual approach against the Textual CoT. The Latent-Visual method outperforms Textual CoT across almost all metrics. We attribute this advantage to the inherent fragility of Textual CoT in circuit analysis: forcing the model to explicitly output long sequences of coordinates for topological paths is highly error-prone, whereas our latent-visual representations preserve structural integrity without relying on strict text generation.
\begin{table*}[h]
\caption{\textbf{Performance comparison of different Chain-of-Thought (CoT) paradigms on topology tasks.} Bold values indicate the best performance. Acc. and EMR denote Accuracy and Exact Match Ratio, respectively.}
\centering
\label{tab:Chain-of-Thought}
\small
\setlength{\tabcolsep}{3pt}
\resizebox{\textwidth}{!}{
    \begin{tabular}{cccccccccccc}
    \toprule
    \multirow{3}{*}{\textbf{Model}} & \multirow{3}{*}{\textbf{Method}} & \multicolumn{3}{c}{Connection Judge.} & \multicolumn{6}{c}{Connection Identification} & \multirow{3}{*}{Avg}\\
    \cmidrule(lr){3-5} \cmidrule(lr){6-11} 
    & & \multicolumn{1}{c}{Easy} & \multicolumn{1}{c}{Medium} & \multicolumn{1}{c}{Hard} & \multicolumn{2}{c}{Easy} & \multicolumn{2}{c}{Medium} & \multicolumn{2}{c}{Hard} & \\
    \cmidrule(lr){3-3} \cmidrule(lr){4-4} \cmidrule(lr){5-5} \cmidrule(lr){6-7} \cmidrule(lr){8-9} \cmidrule(lr){10-11}
    & & Acc.$\uparrow$ & Acc.$\uparrow$ & Acc.$\uparrow$ & F1$\uparrow$ & EMR.$\uparrow$ & F1$\uparrow$ & EMR.$\uparrow$ & F1$\uparrow$ & EMR.$\uparrow$ & \\
    \midrule
    \multirow{2}{*}{GPT-4o} & - & 78.04 & 68.68 & 47.41 & 82.46 & 62.57 & 79.10 & 30.67 & 73.80 & 13.40 & 59.57 \\
    & Visual CoT & 68.44 & 60.94 & 43.68 & 82.54 & 60.48 & 79.66 & 31.73 & 72.18 & 11.00 & 56.74 \\
    \midrule
    \multirow{2}{*}{\begin{tabular}[c]{@{}c@{}}Qwen2.5-VL\\7B\end{tabular}} & Textual CoT & 77.58 & 70.00 & 63.48 & 88.69 & 73.65 & 89.62 & 61.07 & 82.99 & \textbf{29.65} & 70.75 \\
    & Latent-Visual & \textbf{81.42} & \textbf{70.31} & \textbf{64.98} & \textbf{91.44} & \textbf{76.35} & \textbf{91.10} & \textbf{62.67} & \textbf{83.54} & 28.52 & \textbf{72.26} \\ 
    \bottomrule
    \end{tabular}   
}
\end{table*}


\subsection{Sim Weight $\gamma$}
\begin{figure*}[h]
  \centering
  \includegraphics[width=\textwidth]{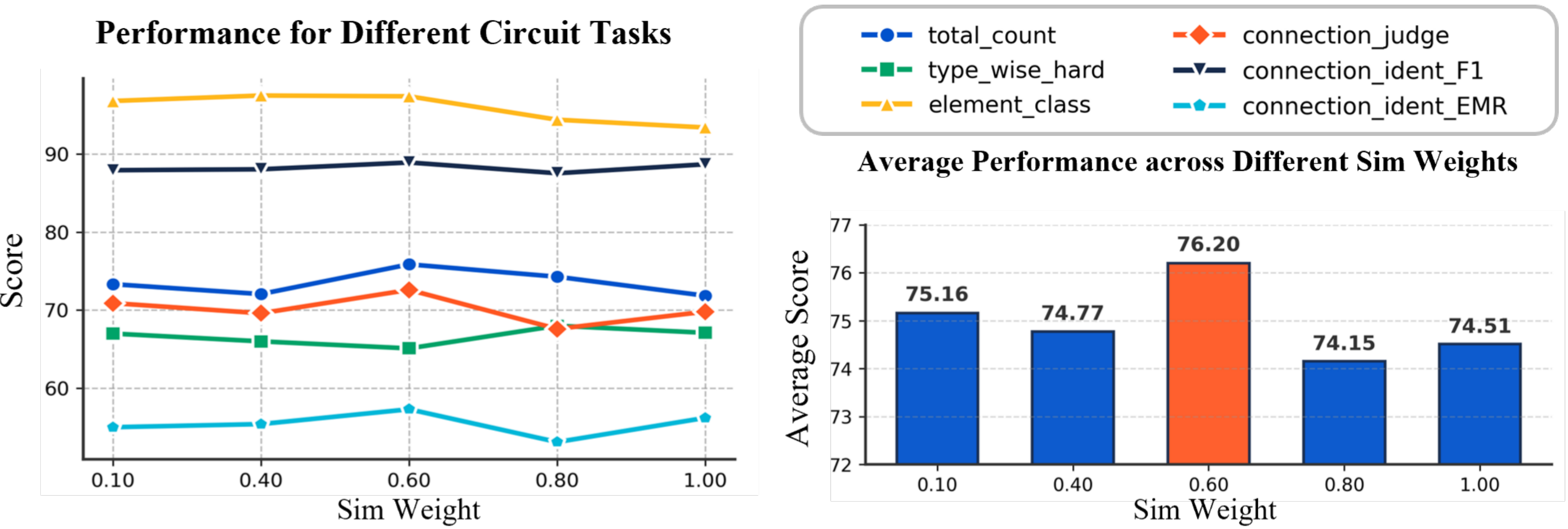} 
  \caption{\textbf{Ablation results of different sim weight $\gamma$ values.} Left: Performance on individual tasks. Right: Average scores across all tasks.}
  \label{fig:sim_weight}
\end{figure*}

To explore the optimal similarity weight $\gamma$, we conducted experiments across various values of $\gamma$ with a fixed latent size of $4$. The overall loss function is defined as $\mathcal{L}_{\text{total}} = \gamma \mathcal{L}_{\text{topo}} + \mathcal{L}_{\text{text}}$. As shown in Figure~\ref{fig:sim_weight}, performance across all tasks peaks when $\gamma$ is set to $0.6$, demonstrating that latent space alignment and text prediction mutually reinforce each other.
\label{sec:Sim Weight}

\subsection{Amsbench}
\label{sec:Amsbench}
In addition to our custom Circuit-MLLM-Bench, we evaluate our approach on AMSBench \cite{shi2025amsbench}. We compare against mainstream general-purpose MLLMs, including closed-source (e.g., GPT-5.1, GPT-4o, Claude-4-Sonnet, Doubao-1.5-vision-pro) and open-source models (e.g., Qwen 2.5 and 3 series, DeepSeek-VL2, GLM-4.5V). We also include Masala CHAI \cite{bhandari2024masala} as a baseline representing methods that explicitly modify images via external tools prior to analysis. 

As shown in \Cref{tab:amsbench}, Circuit-MLLM outperforms all baselines on Type Counting and the topology-focused Connection Identification tasks, demonstrating its strong capability across general circuit tasks. Notably, on Connection Identification, our model achieves a 58\% improvement over its backbone, Qwen-2.5-VL 7B, and an approximate 15\% increase over GPT-4o, underscoring its proficiency in topology analysis. Additionally, we observe that our Circuit-MLLM underperforms models such as GPT5.1 on general circuit problems, specifically the total count category in amsbench. We attribute this performance gap to the fact that the regions of interest for total count tasks typically span the entire image, which renders additional focusing mechanisms in the latent space less effective. 
Conversely, Masala CHAI underperforms the Qwen-2.5-VL 7B backbone across almost all tasks, suggesting that explicit pixel-level modifications on the original image offer limited effectiveness.

\begin{table*}[h]
\caption{\textbf{Performance comparison on Amsbench.} We report Accuracy (Acc.) and Mean Squared Error (MSE) for the Total Counting and Type Counting tasks, and the F1-score (F1) for the Connection Identification task. }
\label{tab:amsbench}
\centering
\setlength{\tabcolsep}{8.pt}
\resizebox{\textwidth}{!}{
\begin{tabular}{llccccc}
    \toprule
    \multicolumn{1}{l}{\multirow{3}{*}{\textbf{Models}}} & \multicolumn{1}{l}{\multirow{3}{*}{\textbf{Size}}} & \multicolumn{5}{c}{\textbf{Amsbench}} \\
    \cmidrule(lr){3-7}
     & & 
    \multicolumn{2}{c}{Total Counting} & \multicolumn{2}{c}{Type Counting} & \multicolumn{1}{c}{Connection Identification}\\
    \cmidrule(lr){3-4} \cmidrule(lr){5-6} \cmidrule(lr){7-7}
     & & 
    Acc.$\uparrow$ & MSE.$\downarrow$ & Acc.$\uparrow$ & MSE.$\downarrow$& F1$\uparrow$\\
    
    \midrule
    GPT-5.1 & / & 35.53 &\textbf{17.96}  &58.26  &10.92 &65.43  \\
    GPT-4o & / & \textbf{51.00} &19.05  &54.00  &28.18    &65.00  \\
    Claude-3.7-Sonnet &/ &36.00  &18.38  &55.00  &24.18   &71.00  \\ 
    Doubao-1.5-vision-pro &/ &24.00  &38.13  &51.00  &24.76  &64.00  \\
    deepseek-vl2 &27B   &13.40 	&76.83 &30.93 	&27.11 	&30.85 \\
    GLM-4.5V &106B     &48.20 	&25.59 		&58.73 	&23.79  &51.27 \\
    \midrule
    \multirow{2}{*}{Qwen3-VL}
    & 8B  &15.87&137.57	&35.40 	&29.26 	&41.27 \\
    & 32B     &27.67 	&28.33 		&54.80 	&19.73 	&46.89 \\
    \midrule
    \multirow{3}{*}{Qwen2.5-VL}  & 7B  &17.07  &84.48  &49.00  &35.49 &47.36  \\
    & 32B &20.13 &	60.21 &50.53 	&31.49 	&47.83  \\
    & 72B &43.00  &19.59   &54.33 &18.59   &52.00  \\ 
    \midrule
    Masala-CHAI & 7B &14.00  &95.93  &45.27   &50.44 &44.79  \\
    \textbf{Circuit-MLLM} & 7B  &36.05  &38.28  &\textbf{60.46} &\textbf{5.64}  & \textbf{75.01}  \\
    \bottomrule
    \end{tabular}    
    }
\end{table*}

\section{Circuit-MLLM-Bench}
\label{sec:Circuit-MLLM-Bench}
Unlike AMSBench, our Circuit-MLLM-Bench strictly evaluates the visual information extraction capabilities of MLLMs on circuit diagrams. To construct this benchmark, we aggregated schematics from four primary non-benchmark datasets: AMSnet \cite{shi2024amsnet}, AnalogGenie \cite{gao2025analoggenie}, Masala-Chai \cite{bhandari2024masala}, and MAPs \cite{zhu2025maps}.

\begin{figure*}[h]
  \centering
  \includegraphics[width=0.95\textwidth]{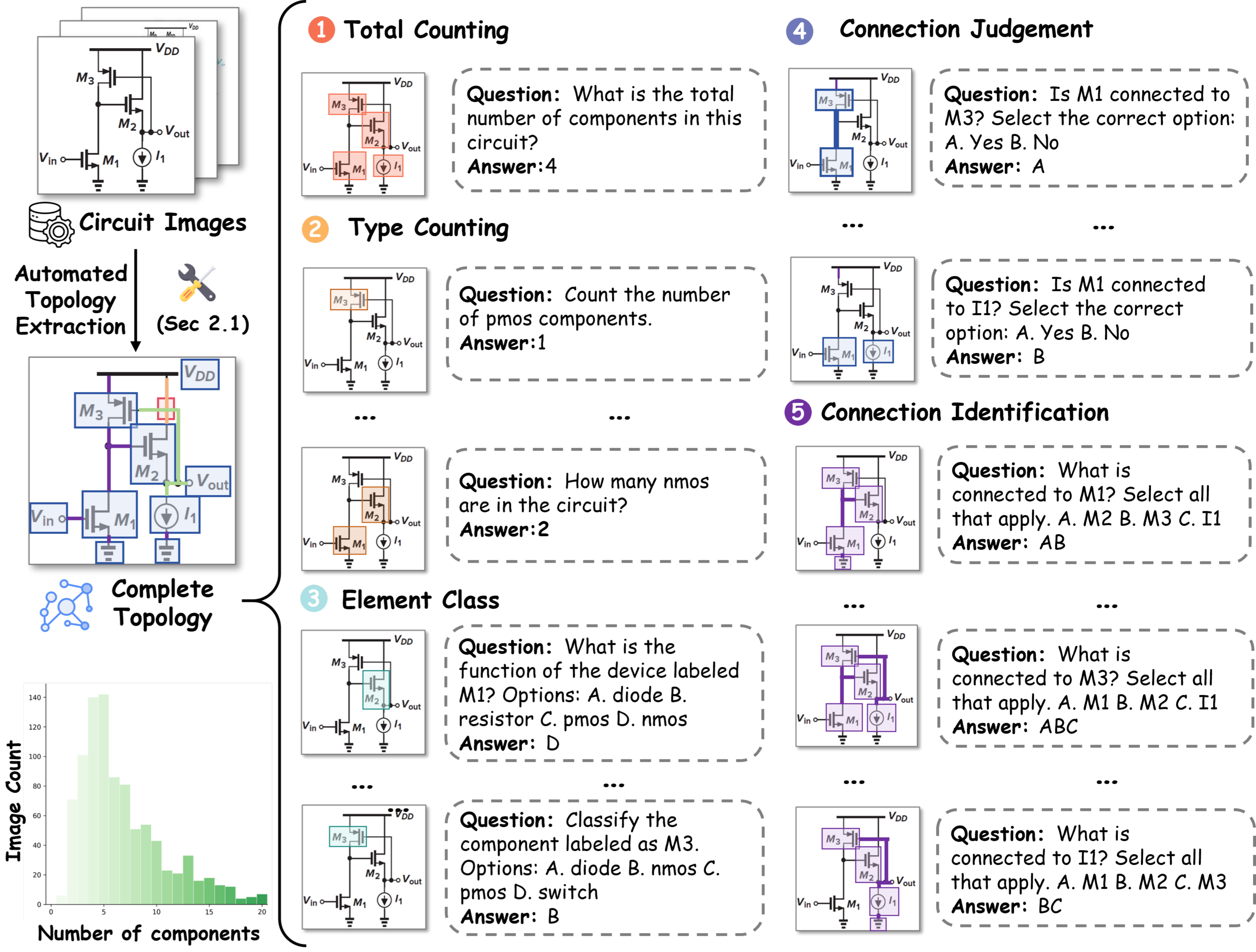} 
  \caption{\textbf{Construction of Circuit-MLLM-Bench.}}
  \label{fig:bench}
\end{figure*}

As illustrated in \Cref{fig:bench}, the collected images are first deduplicated. Subsequently, comprehensive topological data is extracted using the automated pipeline detailed in \Cref{sec:Topology Extraction}. This topology serves as the foundation for synthesizing five distinct task categories: Connection Judge, Connection Identification, Total Count, Type Count, and Element Class. While Total Count yields exactly one query per schematic, the other categories generate multiple instances, resulting in over 20 question-answer (QA) pairs per image.

For rapid and efficient evaluation, the Circuit-MLLM-Bench utilized in this study comprises a curated subset of 1,000 schematics, containing 1,000 QA pairs per task (5,000 QA pairs in total). Note that this pipeline is highly scalable, capable of generating hundreds of thousands of QA pairs if applied to the entire schematic corpus. To ensure rigorous data quality, the generated dataset underwent preliminary automated verification via Qwen-3-VL (32B), followed by meticulous manual review by domain experts.
\section{Case Studies on Topology Analysis with Circuit-MLLM}
\begin{figure*}[h]
  \centering
  \includegraphics[width=0.95\textwidth]{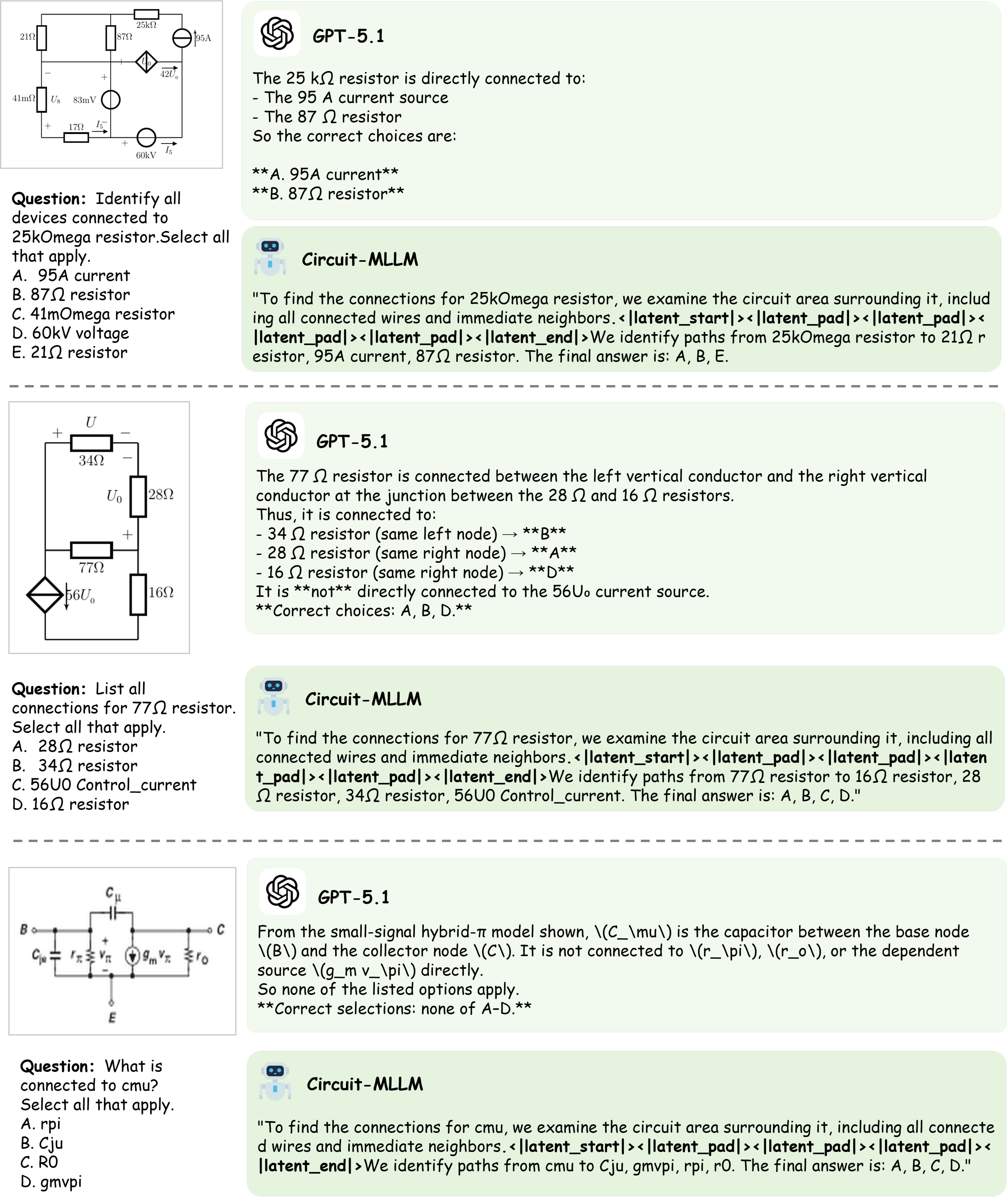} 
  \caption{\textbf{Representative examples of topology analysis.}}
  \label{fig:example}
\end{figure*}

As illustrated in \Cref{fig:example}, in contrast to GPT-5.1, which frequently struggles to interpret complex and densely routed circuit connections, Circuit-MLLM demonstrates superior performance by effectively utilizing its latent-visual reasoning capabilities to accurately identify all interconnected components.
\label{sec:example}
\end{document}